\RequirePackage{fix-cm}
\documentclass[smallextended]{svjour3}       
\smartqed  
\usepackage{graphicx}
\usepackage{algorithm}
\usepackage{algpseudocode}

\usepackage[]{fontenc}
\usepackage[latin1]{inputenc}
\usepackage{verbatim}
\usepackage{amsmath}
\usepackage{amssymb}
\usepackage{dsfont}
\usepackage{amsfonts}
\usepackage{mathrsfs}
\usepackage{graphicx}
\usepackage{color}
\usepackage{ulem}
\usepackage{xspace,bbm,epic,eepic}
\usepackage{pgfplots}
\usepackage{setspace}
\usepackage{xspace}

\usepackage{ stmaryrd }
\usepackage{graphicx}
\usepackage{lineno}
\usepackage{color}
\usepackage{url}
\usepackage{graphicx}
\usepackage{ stmaryrd }
\usepackage{algorithm}
\usepackage{algpseudocode}

\usepackage{amsmath,amssymb,color} 

\newcommand{\Rr}{\mathds{R}}
\newcommand{\Cr}{\mathds{C}}

\newcommand{\R}{P}

\newcommand{\Q}{Q}

\newcommand{\XXX}{\mathbf{X}} 
\newcommand{\YYY}{\mathbf{Y}} 
\newcommand{\AAA}{{\mathbf{X}}} 
\newcommand{\BBB}{{\mathbf{Y}}} 
\newcommand{\ZZZ}{\mathbf{Z}}

\newcommand{\eg}{\textit{e.g.}, }
\newcommand{\ie}{\textit{i.e.}, }

\DeclareMathOperator*{\argmin}{arg\,min}

\begin{document}

\title{State-Of-The-Art Algorithms For Low-Rank Dynamic Mode Decomposition.
}


\author{Patrick H\'eas          \and
        C\'edric Herzet  
}

\institute{INRIA Centre Rennes - Bretagne Atlantique \&
		IRMAR - UMR CNRS 6625,\\
		campus universitaire de Beaulieu, 35042 Rennes, France.\\
              \email{patrick.heas@inria.fr}           
          }

\date{}

\maketitle

\begin{abstract}
This  technical note reviews sate-of-the-art  algorithms for linear  approximation of high-dimensional dynamical systems using low-rank dynamic mode decomposition (DMD). While repeating several parts of the article  \cite{HeasHerzet2021}, this work provides useful complementary details to build up an overall picture of state-of-the-art methods.  
\end{abstract}

\section{Introduction} \vspace{-0.15cm}

\subsection{Context}\label{sec:context}
The numerical discretization of a  partial differential equation parametrized by its initial condition often leads to a very high dimensional system  {of the form:} \vspace{-0.cm}
\begin{align}\label{eq:model_init} 
 \left\{\begin{aligned}
& x_{t}(\theta)= f_t(x_{t-1}(\theta)) \\
&x_1(\theta)={\theta}
\end{aligned}\right. ,
\quad t = 2 ,\ldots, T,\vspace{-0.cm}
\end{align} 
\noindent
 {where}  $x_t(\theta)\in \Rr^n$ is the state variable,   $f_t:\Rr^n \to \Rr^n$, and  $\theta \in \Rr^n$ denotes an initial condition. 
In some context, \eg for uncertainty quantification purposes, one is interested by  computing  a set of  trajectories corresponding to different initial conditions $\theta \in \Theta \subset \Rr^n$. This may constitute an intractable task due to the  high dimensionality of the space embedding the trajectories. For instance, in the case where $f_t$ is linear, the  complexity   required to compute a trajectory of  model \eqref{eq:model_init}   scales in $\mathcal{O}(Tn^2)$, which is prohibitive for large values of  $n$ or  $T$.

 To deal with these large values, reduced models approximate  the trajectories of the system for a range of regimes determined by a set of initial conditions~\cite{2015arXiv150206797C}. 
 A common assumption is that the trajectories of interest are well approximated in a low-dimensional subspace of $\Rr^n$. In this spirit, many  tractable approximations of model \eqref{eq:model_init}  have been proposed, in particular the well-known  {\it Petrov-Galerkin projection}~\cite{quarteroni2015reduced}. 
However, these methods require   the knowledge of the equations ruling the high-dimensional system. 

Alternatively, there exist   data-driven approaches. In particular, linear inverse modeling~\cite{penland1993prediction}, 
 principal oscillating patterns~\cite{Hasselmann88}, or more recently, dynamic mode decomposition (DMD)~\cite{Chen12,2016Dawson,hemati2017biasing,Jovanovic12,kutz2016dynamic,Schmid10,Tu2014391} propose to  approximate the unknown function $f_t$ by a linear and low-rank operator. This linear framework  has been extended to  quadratic approximations of $f_t$ in~\cite{CuiMarzoukWillcox2014}.  Although  linear approximations are in appearance restrictive,  they have  recently sparked a new surge of interest because they are at the core of the so-called extended DMD or kernel-based  DMD     \cite{budivsic2012applied,li2017extended,williams2015data,williams2014kernel,2017arXiv170806850Y}.  The latter   decompositions   characterize accurately non-linear behaviours under certain conditions~\cite{klus2015numerical}.


  Reduced models based on  low-rank linear approximations substitute  function $f_t$ by a matrix   $\hat A_k \in \Rr^{n \times n}$ with $r={\textrm{rank}(\hat {A}_k)} \le n$   as
\begin{align}\label{eq:model_koopman_approx} 
 \left\{\begin{aligned}
& \tilde x_{t}(\theta)=  \hat A_k \tilde  x_{t-1}(\theta),\quad t=2,\ldots,T,  \\
&\tilde x_1(\theta)={\theta},
\end{aligned}\right. \vspace{-0.cm}
\end{align} 
where  $\{\tilde x_{t}(\theta)\}_{t=1}^T$ denotes an approximation of the trajectory $\{x_t(\theta)\}_{t=1}^T$ of system \eqref{eq:model_init}. 
The complexity for the evaluation of a trajectory approximation with \eqref{eq:model_koopman_approx} will be refered to as {\it on-line complexity}. A low on-line complexity is obtained by exploiting  the low rank  of matrix $\hat A_k$.  A  scaling  in  $\mathcal{O}(Tr^2+rn)$ is reached if the reduced model is parametrized by  matrices  $R,\,L \in \Cr^{n \times r}$ and $S \in \Cr^{r \times r}$  such that  trajectories of \eqref{eq:model_koopman_approx} correspond to  the  recursion
\begin{equation}\label{eq:genericROM0}
 \left\{\begin{aligned}
& \tilde x_{t}(\theta) =  R  z_t,  \quad t=2,\ldots,T,\\
& z_t= S  z_{t-1},  \,\,\,\quad t=3,\ldots,T,\\
&z_2=L^\intercal \theta. 
\end{aligned}\right.
\end{equation}
The equivalence of  systems    \eqref{eq:model_koopman_approx} and \eqref{eq:genericROM0} is obtained for   $T\ge 2$ by setting  
$\hat A_k^{T-1}=R S^{T-2}L^\intercal$. 
In particular, consider a factorization of the form 
\begin{align}\label{eq:solFactor}
\hat A_k=PQ^\intercal \quad \textrm{with}\quad P,Q \in \Rr^{n \times r}. 
\end{align}
 This factorization is always possible by computing the singular value decomposition (SVD)  ${\hat A_k}=U_{\hat A_k}\Sigma_{\hat A_k} V_{\hat A_k}^\intercal$ and identifying $P=U_{\hat A_k}$ and $Q^\intercal =\Sigma_{\hat A_k} V_{\hat A_k}^\intercal$. 
 Factorization~\eqref{eq:solFactor} implies  that trajectories of \eqref{eq:model_koopman_approx} are  obtained with system~\eqref{eq:genericROM0}  setting $R=\R$, $L=\Q$ and $S=  \Q^\intercal  \R$.
 Another factorization of interest relies on the eigenvalue decomposition (EVD) 
\begin{align}\label{eq:factorEVD}
\hat A_k = D \Lambda D^{-1},\quad \textrm{with}\quad D,\Lambda \in \Cr^{n \times n},
\end{align}
where   $\Lambda$ is a Jordan-block matrix \cite{golub2013matrix}  of rank  $r.$ 
Using the ``economy size''  EVD yields a system of the form of \eqref{eq:genericROM0}. Indeed, it is obtained by making  the identification 
  $L=(\xi_1\cdots \xi_r) $ and  $R=(\zeta_1\cdots \zeta_r)$, where $\xi_i\in \Cr^n$ and $\zeta_i\in \Cr^n$ are the $i$-th left and right eigenvectors of $\hat A_k$ (equivalently the $i$-th column of $(D^{-1})^\intercal$ and $D$), and identifying $S$ to the first $r \times r$ block of $\Lambda$ multiplied by $L^\intercal$.

The on-line complexity to compute this recursion is still  $\mathcal{O}(Tr^2+rn)$.  But  assuming that $\hat {A}_k$ is diagonalizable\footnote{Diagonalizability is guaranteed if all the non-zero eigenvalues are distinct. However, this condition is  only sufficient and the class of diagonalizable matrices is  larger \cite{Horn12}.}, we have $S =\textrm{diag}(\lambda_1,\cdots, \lambda_r)$ and system~\eqref{eq:genericROM0} becomes
\begin{align}\label{eq:koopman1}
\left\{\begin{aligned}
 \tilde x_{t}(\theta)&= \sum_{i=1}^{r}\zeta_i \nu_{i,t},\\
\nu_{i,t}& =  \lambda_i^{t-1} \xi_i^\intercal  \theta, \quad \textrm{for} \quad i=1,\ldots, \textrm{rank}(\hat {A}_k)
\end{aligned}\right. , \quad t=2,\ldots,T,\vspace{-0.cm}
\end{align} 
where  $\lambda_i \in \Cr$  is the $i$-th  (non-zero) eigenvalue of $\hat {A}_k$.  This reduced-model possesses  
 a very desirable  on-line complexity of  $\mathcal{O}(rn)$,  
\ie  linear in the ambient dimension $n$, linear in the reduced-model intrinsic dimension $r$ and independent of the trajectory length~$T$. \\

The key of reduced modeling  is to find a ``good'' tradeoff between  the  on-line complexity and the  accuracy of the approximation. As shown previously, the  low on-line computational effort is obtained by a proper factorization of the low-rank matrix $\hat A_k$. Thus,  in an off-line stage, it remains  to {\it i)}~search $\hat A_k$ within the family of low-rank matrices which yields the ``best''  approximation~\eqref{eq:model_koopman_approx}, {\it ii)}~compute the SVD or EVD based factorization  of   $\hat A_k$.   We will refer to  the computational cost associated to these two steps as {\it off-line complexity}.


%
%

A standard choice is to select $\hat A_k$ inducing   the best  trajectory approximation in the $\ell_2$-norm sense, for initial conditions in the set  $\Theta \subset \Rr^n$:  matrix  $\hat A_k$ in \eqref{eq:model_koopman_approx} targets  the solution of the following minimization problem for some given $k \le n $:\vspace{-0.2cm}
\begin{align}\label{eq:target}
\argmin_{A:\textrm{rank}(A)\le k}  \int_{\theta \in \Theta} \sum_{t=2}^T \| x_{t}(\theta) - A^{t-1} \theta\|^2_2, 
\end{align}
where $\|\cdot\|_2$ denotes the $\ell_2$-norm. 
Since we focus on  {\it data-driven} approaches, we  assume that  we do not know the exact form of $f_t$ in~\eqref{eq:model_init}
 and we only have access to a set of representative trajectories $\{x_t(\theta_i)\}_{t=1}^T$, $i=1,...,N$ so-called \textit{snapshots}, obtained by running the high-dimensional system for $N$ different initial conditions $\{\theta_i\}_{i=1}^N$ in the set $\Theta$.   
Using these snapshots, we consider a discretized version of~\eqref{eq:target}, which corresponds to the constrained optimization problem studied  in \cite{Chen12,Jovanovic12,wynn2013optimal}: matrix $\hat A_k$ now targets  the solution 
\begin{align}\label{eq:prob} 
A_k^\star \in &\argmin_{A:\textrm{rank}(A)\le k}  \sum_{i=1}^{N}  \sum_{t=2}^{T} \| x_{t}(\theta_i) - A   x_{t-1}(\theta_i)\|^2_2, 
\end{align} 
 where we have substituted   $A^{t-1} \theta_i$ in \eqref{eq:target} by   $A   x_{t-1}(\theta_i)$ and where we have approximated the integral by an empirical average over the snapshots. 

 Problem~\eqref{eq:prob} is non-convex due to the presence of the rank constraint ``$\textrm{rank}(A)\le k$''. As consequence, it has  been considered as intractable in several contributions of the litterature and numerous procedures have been proposed to approximate its solution (see next section). 
The work ~\cite{HeasHerzet2021} shows that problem \eqref{eq:prob} is in fact tractable and admits a closed-form solution which can be evaluated in polynomial-time. \vspace{-0.15cm}
%


%
\subsection{Problem Statement}

The off-line construction of reduced models of the form of~\eqref{eq:genericROM0} focuses on the following  questions:
\begin{enumerate}
  \item Can we compute a solution of problem \eqref{eq:prob}  in polynomial time?
  \item How to compute efficiently a factorization of this solution, and in particular its EVD?
\end{enumerate}
Let us make some correspondences with the terminology used in the DMD literature~\cite{Chen12,2016Dawson,hemati2017biasing,Jovanovic12,kutz2016dynamic,Schmid10,Tu2014391} in order to reformulate these two questions in the jargon used in this community. 
The   { ``low-rank DMD''}  of system \eqref{eq:model_init} refers to the EVD of the solution $A_k^\star$ of problem \eqref{eq:prob}, or equivalently to the parameters of reduced model~\eqref{eq:koopman1}  in the case where $\hat A_k=A_k^\star$ is diagonalizable.\footnote{The ``DMD''  of system \eqref{eq:model_init} refers to the EVD of the solution of problem \eqref{eq:prob} without the low-rank constraint.}  
Using this terminology, the two above questions can be summarized summarized as follows: can we compute exactly and with a polynomial complexity the low-rank DMD  of system \eqref{eq:model_init}? 
The answer to this question is positive as proved in~\cite{HeasHerzet2021}. \\

\textbf{Solver for problem \eqref{eq:prob}.}\,
In the last decade, there has been a surge of interest for  low-rank solutions of linear matrix equations, see \eg~\cite{fazel2002matrix,jain2010guaranteed,lee2009guaranteed,lee2010admira,mishra2013low,recht2010guaranteed}.  This  class of  problems includes \eqref{eq:prob} as an important  particular  case. Problems in this class  are always non-convex due to the rank constraint and  computing their solutions in polynomial time  is often out of reach. Nevertheless,   certain instances of these problems with  very special structures admit closed-form solutions \cite{eckart1936approximation,mesbahi1997rank,parrilo2000cone}. The work \cite{HeasHerzet2021} shows that \eqref{eq:prob} belongs to this class of problems and provide a closed-form solution which can be computed in polynomial time. Prior to this work, many authors have proposed tractable procedures to compute approximations of the solution to problem  \eqref{eq:prob}~ \cite{Chen12,Jovanovic12,li2017extended,Tu2014391,wynn2013optimal,2017arXiv170806850Y} or to related problems~\cite{hemati2017biasing}. We review these contributions in Section~\ref{sec:approxLDMD} and discuss their complexity.\vspace{-0.15cm}\\

\textbf{Factorization of the solution.}
The second problem concerns the computation of the factorization of the form \eqref{eq:solFactor} or \eqref{eq:factorEVD} of the solution $A_k^\star  \in \Rr^{n \times n}$. 
A brute-force computation of a factorization of a matrix in $\Rr^{n \times n}$, in particular an EVD, is  prohibitive for  large values of $n$. The paper \cite{HeasHerzet2021} proposes  low-complexity algorithms  computing such factoization of $A_k^\star$. This follows the line and extends previous works~\cite{Jovanovic12,Tu2014391,williams2014kernel}, as detailed in Section \ref{sec:etatArt2}.\vspace{-0.15cm}\\

The following sections  
provide a  review on techniques for approximating and factorizing the solution of problem \eqref{eq:prob}. \vspace{-0.25cm}
 \section{Notations}\label{sec:notations}\vspace{-0.05cm}
 All along the paper, we   make extensive use    of the  economy-size SVD  of a matrix  $M\in \Rr^{p \times q }$ with $p\ge q$: $M=U_M\Sigma_M V_M^\intercal $ with $U_M\in \Rr^{p \times q }$, $V_M\in \Rr^{ q \times  q}$ and $\Sigma_M\in \Rr^{q  \times q }$ so that $U_M^\intercal U_M=V_M^\intercal V_M=I_q$ and $\Sigma_M $ is diagonal, where the  upper script~$\cdot^\intercal $  refers to the transpose and $I_q$  denotes the $q$-dimensional identity matrix. 
 The columns of matrices $U_M$ and $V_M$ are  denoted $U_M=(u_M^1 \cdots u_M^q)$ and $V_M=(v_M^1 \cdots v_M^q)$ while  $\Sigma_M  =\textrm{diag}( \sigma_{M,1},  \cdots,  \sigma_{M,q})$  with $\sigma_{M,i} \ge \sigma_{M,i+1}$ for $i=1,\ldots, q-1$. The  Moore-Penrose pseudo-inverse of matrix $M$ is then  defined as $M^{\dagger}=V_M\Sigma^{\dagger}_M U_M^\intercal $, where 
 $\Sigma^{\dagger}_M=\textrm{diag}( \sigma_{M,1}^{\dagger}, \cdots , \sigma_{M,q}^{\dagger})$ with
 $$ 
 \sigma_{M,i}^{\dagger}=  \left\{\begin{aligned}
&\sigma_{M,i}^{-1}\quad \textrm{if}\quad  \sigma_{M,i} > 0\\
&0\quad\quad\,\,\,\textrm{otherwise}
\end{aligned}\right. .\vspace{-0.cm}\\
$$
The orthogonal projector onto the span of the columns (resp. of the rows) of matrix $M$ is denoted by $\mathbb{P}_{M}=M M^\dagger=U_M\Sigma_M\Sigma_M^\dagger U_M^\intercal $ (resp. $\mathbb{P}_{M^\intercal}=M^\dagger M=V_M \Sigma_M^\dagger \Sigma_M V_{M}^\intercal$) \cite{golub2013matrix}. 

We also introduce additional notations to derive a matrix formulation of the low-rank estimation problem \eqref{eq:prob}. We gather  consecutive elements of the $i$-th snapshot trajectory between time $t_1$ and $t_2$   in matrix $X_{t_1:t_2}^{(i)}~= ~(x_{t_1}(\theta_i) \cdots x_{t_2}(\theta_i))$ and form large matrices $ \XXX, \YYY \in \Rr^{n \times  m} $ with $m=N(T-1)$  as 
 $$\XXX = (X^{(1)}_{1:T-1} \cdots  X^{(N)}_{1:T-1})  \quad \textrm{and} \quad 
 \YYY= ( X^{(1)}_{2:T} \cdots X^{(N)}_{2:T}).  $$
 In order to be consistent with the SVD definition and to keep the presentation as simple as possible, this work  assumes that $m\leq n$. However, all the result presented in this work can be extended without any difficulty to the case where  $m> n$ by using an alternative definition of the SVD.\vspace{-0.25cm}
		


\section{Sub-Optimal Solutions}\label{sec:stateArt}

We begin by  presenting state-of-the-art methods solving  approximatively the low-rank minimisation problem \eqref{eq:prob}.  In a second part, we  make an overview of state-of-the-art  algorithms  computing   factorisations of these approximated solutions of the form of  \eqref{eq:solFactor} or \eqref{eq:factorEVD}.

 \subsection{Tractable Approximations to Problem~\eqref{eq:prob}}\label{sec:approxLDMD}
Using   the notations introduced in Section~\ref{sec:notations}, problem~\eqref{eq:prob} can be rewritten as
		\begin{align}\label{eq:prob1} 
		A_k^\star \in &\argmin_{A:\textrm{rank}(A)\le k} \|\BBB -A \AAA \|^2_F,
		\end{align} 
where $\|\cdot\|_F$ refers to the Frobenius norm.

   \subsubsection{Truncation of the Unconstrained Solution} \label{sec:trunc} ~
 A first approximation consists in removing the low-rank constraint in problem~\eqref{eq:prob1}. 
As pointed out by {\it Tu et al.} in \cite{Tu2014391},  the problem then  boils down to a  least-squares  problem 
  		\begin{align}\label{eq:prob1_unconst} 
		\argmin_{A} \|\BBB -A \AAA \|^2_F,
		\end{align} 
		 admitting the closed-form solution $\BBB\AAA^{\dagger}$.
Matrix $\BBB\AAA^{\dagger}$ also solves  the constrained  problem \eqref{eq:prob1} in the case where  $k\ge m$ and in particular  for $k=m$, \ie \vspace{-0.15cm}
\begin{align}\label{eq:exactDMD}
A^\star_m=\BBB\AAA^{\dagger}.
\end{align}
This solution relies on the  SVD of $\XXX$: $A^\star_m=\BBB V_{\AAA}\Sigma_{\AAA}^{\dagger}U_{\AAA}^\intercal,$ which is computed with a complexity of $\mathcal{O}(m^2(m+n))$~\cite{golub2013matrix}.  An approximation of the solution of~\eqref{eq:prob1} satisfying the low-rank constraint $\textrm{rank}(A)\le k$ with  $k < m$ is then  obtained by a truncation of the SVD or the EVD of $A^\star_m$ using $k$ terms. \\
 \subsubsection{Approximation by low-rank projected DMD}\label{sec:LowRankProj} ~
 
 The so-called {\it ``projected  DMD''} proposed  by {\it Schmid} in~\cite{Schmid10} is a low-dimensional approximation of $A^\star_m$. This approximation is also used by {\it Jovanovic et al.}  in order to approximate $A^\star_k$ for $k<m$~\cite{Jovanovic12}.  Similar approximations are used to compute the so-called {\it ``optimized DMD''} in \cite{Chen12} or  {\it ``optimal mode decomposition''} in \cite{wynn2013optimal}.These approximations assume that  columns of matrix $A\AAA$ are in the span of $\AAA $.   
 This assumption is  formalised   in \cite{Schmid10}  as the existence of $A^c\in \Rr^{ m \times m}$, the so-called  \textit{``companion matrix''} of some matrix $A$ parametrised by $m$ coefficients,\footnote{The exact definition  of  the ``companion matrix'' $A^c$ considered by {\it Schmid} is as follows: 
 \begin{align}\label{eq:companion2}
 A^c=
 \begin{pmatrix}
 0 & &  & & \alpha_1\\
 1 &0&  && \alpha_2\\
   &\ddots & \ddots& &\vdots\\
 & & 1& 0& \alpha_{m-1}\\
 && &1& \alpha_m  
  \end{pmatrix} \in \Rr^{m \times m}.
   \end{align}
It depends on  the  $m$ coefficients $\{\alpha_i\}_{i=1}^m$, see details in \cite{Schmid10}.
 } such that 
\begin{align}
A \AAA=\AAA  A^c.\label{eq:companion}
  \end{align} 
Under this assumption, we obtain from~\eqref{eq:companion}  a low-dimensional representation  of $A$ in the span of $U_{\AAA}$,  
\begin{align}\label{eq:DMDassumption}
 U_{\AAA}^\intercal AU_{\AAA}=\tilde A^c,
\end{align}
where $ \tilde A^c=\Sigma_{\AAA} V_{\AAA}^\intercal A^cV_{\AAA}\Sigma_{\AAA}^{\dagger}\in \Rr^{ m \times  m}$.
 {\it Jovanovic et al.}  then obtain an approximation of $A^\star_k$   by plugging~\eqref{eq:companion} in problem~\eqref{eq:prob1} and  minimising the resulting cost with respect to $A^c$~\cite{Jovanovic12}. 
Using the invariance of the Frobenius norm to unitary transforms, this approximation of $A^\star_k$  can be rewritten as the solution of 
\begin{align}\label{eq:DMDSVD}
\argmin_{ \tilde A^c: \textrm{rank}(\tilde A^c\Sigma_{\AAA})\le k} \|U_{\AAA}^\intercal \BBB V_{\AAA} - \tilde A^c \Sigma_{\AAA}\|^2_F.
\end{align}
Assuming  $\AAA$ is full-rank, the solution is   given by the Eckart-Young theorem~\cite{eckart1936approximation}: the solution is the SVD representation of matrix $B=U_{\AAA}^\intercal \BBB V_{\AAA}$  truncated to $k$ terms  multiplied by matrix $\Sigma_{\AAA}^{\dagger}$.  Denoting by  $\tilde B$ this truncated decomposition, we finally obtain  the following  approximation of~\eqref{eq:prob1}  
 \begin{align}\label{eq:projDMD}
A_k^\star\approx U_{\AAA}\tilde B \Sigma_{\AAA}^{\dagger}U_{\AAA}^\intercal .
 \end{align}

 
This method relies on the SVD of   $\AAA\in \Rr^{n \times m}$ and $B\in \Rr^{m \times m}$ and thus involves a complexity of  $\mathcal{O}(m^2(m+n))$~\cite{golub2013matrix}. 

  \subsubsection{Approximation by  Sparse DMD}\label{sec:sparse} ~

{\it Jovanovic et al.} also propose  in \cite{Jovanovic12} a two-stage approach  they call {\it ``sparse DMD''}. It  consists in solving two  independent problems. The first stage computes the EVD of the approximated solution~\eqref{eq:projDMD} for $k=m$. This first stage yields  eigen-vectors $\zeta_i,$ for $ i=1, \ldots, m$. In a second stage, the authors assume that a linear combination of $k$ out of the $m$  eigen-vectors  approximates  accurately the data.  This assumption serves to design a relaxed convex optimisation problem using an $\ell_1$-norm penalisation.\footnote{The penalisation parameter must be adjusted to induce $m-k$ coefficients nearly equal to zero.} Solving this problem, they obtain  $k$ eigen-vectors and their associated coefficients.  Note that the sparse DMD approximation has an error norm always greater or equal than the one induced by an approximation by low-rank projected DMD.\footnote{ By decomposing the error in two orthogonal components and by using the invariance of the Frobenius norm to unitary transforms, for any $A$ satisfying    \eqref{eq:companion}, we have 
$ 
\|\BBB-A \AAA  \|^2_F=\|\BBB-\AAA A^c \|^2_F=
 \|U_{\AAA}^\intercal \BBB V_{\AAA}- A^c \Sigma_{\AAA}\|^2_F  + \|( U_{\AAA}^\perp)^\intercal \BBB \|^2_F,  
 $ 
where the columns of $U_{\AAA}^\perp$ contain the $n-m$ vectors orthogonal to $ U_{\AAA}$.
 Taking the minimum  over the set of low-rank companion matrices, we  construct a lower bound on the error norm
\begin{align*}
\min_{\tilde A^c : \textrm{rank}(\tilde A^c\Sigma_{\AAA})\le k} \|U_{\AAA}^\intercal \BBB V_{\AAA}-\tilde A^c \Sigma_{\AAA}\|^2_F + \|( U_{\AAA}^\perp)^\intercal \BBB \|^2_F \le  \|\BBB- A \AAA\|^2_F,
\end{align*}
 for any $A$ satisfying  assumption \eqref{eq:companion}. The lower bound is reached by definition if $A$ is the approximated solution~\eqref{eq:projDMD}. The sparse DMD approximation is built upon assumption \eqref{eq:companion} and thus  has an error norm above or equal  this bound.}

This method relies on   the resolution of an $\ell_1$-norm minimisation of a cost function built using  the EVD of  approximation~\eqref{eq:projDMD} for $k=m$, which is easily  deduced from the EVD of $B \Sigma_{\AAA}^{\dagger} \in \Rr^{m \times m}$. The overall complexity is   $\mathcal{O}(m^2(m+n))$. 

\subsubsection{Approximation by  Total-Least-Square DMD}\label{sec:tls} ~
To ease the presentation, we  reformulate the  total-least-square (TLS) DMD problem studied by {\it Hemati et al.}~\cite{hemati2017biasing}.  Let us define the projector  $\mathbf{V}^k_{\mathbf{K}}(\mathbf{V}^k_{\mathbf{K}})^\intercal$ where columns of $\mathbf{V}^k_{\mathbf{K}} \in \mathbb{R}^{m \times k}$ are the right singular vectors  associated  to the  $k$ largest  singular values of  matrix  $\mathbf{K}  =\begin{bmatrix} \AAA \\ \BBB \end{bmatrix}\in \Rr^{2n \times m}$. The approximation introduced in {\it Hemati et al.}  can be formulated as the solution of the following unconstrained convex optimization problem 
\begin{align}\label{eq:probHemati}
\argmin_{A\in \Rr^{n \times n}} \| \BBB' - A \AAA' \|^2_F,
\end{align}
where $\AAA'= \AAA \mathbf{V}^k_{\mathbf{K}}(\mathbf{V}^k_{\mathbf{K}})^\intercal$ and $ \BBB'= \BBB  \mathbf{V}^k_{\mathbf{K}}(\mathbf{V}^k_{\mathbf{K}})^\intercal$. The solution  of the least square problem \eqref{eq:probHemati} 
\begin{align}\label{eq:probHematiSol}
A_k^\star\approx \BBB\mathbf{V}^k_{\mathbf{K}}(\mathbf{V}^k_{\mathbf{K}})^\intercal\AAA^\dagger,
\end{align}
may constitute an approximation of the solution of the  problem of interest, although
the unconstrained problem~\eqref{eq:probHemati} is  intrinsically different from the low-rank approximation problem  \eqref{eq:prob1}. An analytical example provided in Appendix~\ref{app:comp} highlights how the  solutions of these two different problems differ.  This method relies on the SVD of   $\mathbf{K}\in \Rr^{2n \times m}$ and $\XXX \in \Rr^{n \times m}$ and thus involves a complexity of  $\mathcal{O}(m^2(m+n))$. 

\subsubsection{Approximation by Solving Regularised Problems}\label{sec:convex} ~
Some works propose to approximate~\eqref{eq:prob1} by a  regularized version of the unconstrained  problem~\eqref{eq:prob1_unconst}, using Tikhonov penalization \cite{li2017extended}  or  penalization  enforcing structured sparsity \cite{2017arXiv170806850Y}. However, these  choices of regularizers do not guarantee in general  that the solution is low-rank. In contrast,  the  solution of \eqref{eq:prob1} may under certain theoretical conditions~\cite{lee2010admira,jain2010guaranteed}  be recovered by  the following quadratic  program   
\begin{align}\label{eq:probConvexRelas} 
 A_k^\star &\approx\argmin_{A\in \Rr^{n \times n}} \|\BBB -A \AAA \|^2_F+ \alpha_k \| A \|_{*},\nonumber \\
 &=  \argmin_{A\in \Rr^{n \times n}} \min_{B\in \Rr^{n \times n}} \|\BBB -A \AAA \|^2_F+ \alpha_k \| B \|_{*} \quad \textrm{s.t.} \quad A=B
\end{align} 
where $\| \cdot \|_{*}$  refers to the nuclear norm (or trace norm) of the matrix, \ie  the sum of its singular values.
In optimization problem~\eqref{eq:probConvexRelas}, $\alpha_k \in \Rr_+$ represents an appropriate regularization parameter determining the rank $k$ of the solution. 
Program~\eqref{eq:probConvexRelas} is a convex optimization problem \cite{mishra2013low} which can be efficiently solved using modern optimization techniques, such as the alternate directions of multipliers method (ADMM)~\cite{bertsekas1995nonlinear}. 
The algorithms solving \eqref{eq:probConvexRelas} typically  involve per iteration a complexity  of  $\mathcal{O}(m(m^2+n^2))$. 
\subsection{Factorization of  Approximations of   $ A_k^\star$,}\label{sec:etatArt2}

In this section, we provide an overview of some state-of-the-art methods to compute 
factorizations of the form of~\eqref{eq:solFactor} or~\eqref{eq:factorEVD} 
for the approximations of $ A_k^\star$ presented above.

We first note that a brute-force computation of the SVD or EVD of a matrix in $\Rr^{n \times n}$ leads in general to a prohibitive computational cost since it requires a  complexity of  $\mathcal{O}(n^3)$. Hopefully,  the factorizations \eqref{eq:solFactor} or~\eqref{eq:factorEVD}  are computable with a  complexity of  $\mathcal{O}(m^2(m+n)) $,  in most cases mentioned above.
 
In particular,  in the case of low-rank projected DMD, a straightforward  factorization of the form of~\eqref{eq:solFactor}  is $P=U_\AAA$ and $Q^\intercal = \tilde B \Sigma_{\AAA}^{\dagger}U_{\AAA}^\intercal$. In the case of  sparse DMD, the latter factorization holds by substituting   $\tilde B$ by the ``sparse'' approximation of $B$. Another straightforward  factorization of the form of~\eqref{eq:solFactor} is  intrinsic to the ADMM procedure, which uses an SVD  to compute the regularized solution.

Concerning EVD factorization,  in the case of the truncated approach, {\it Tu et al.} propose an algorithm scaling in   $\mathcal{O}(m^2(m+n))$~\cite{Tu2014391}. 
 In the context of low-rank  projected DMD or sparse DMD, {\it Jovanovic et al.} propose a  procedure of analogous  complexity, which approximates the first $m$ eigenvectors, and then estimate the related eigenvalues by solving a convex optimization problem \cite{Jovanovic12}. In the case of TLS DMD, the diagonalization of a certain matrix in $\Rr^{m \times m}$ suffices to obtain the sought EVD factorization. \\ 

\section{ Optimal Solution in Polynomial Time}\label{sec:contrib}

In this section, we provide the closed-form solution to problem~\eqref{eq:prob1} proposed in \cite{HeasHerzet2021}.  Algorithms are then proposed to compute and factorize this solution in the form of~\eqref{eq:solFactor} or~\eqref{eq:factorEVD}. \vspace{-0.15cm}

\subsection{Closed-Form Solution to~\eqref{eq:prob1}}\label{sec:closedSol}

 Let the columns of matrix  ${U}_{\ZZZ,k} =\begin{pmatrix} u^1_\ZZZ& \cdots &u^k_\ZZZ  \end{pmatrix}\in \Rr^{n \times k}$ be the  left singular vectors $\{u_\ZZZ^i\}_{i=1}^k$ associated  to the  $k$ largest  singular values of  matrix  \begin{align}\label{eq:defZZZ} \ZZZ= 
 \BBB \mathbb{P}_ {\AAA^\intercal}\in \Rr^{n \times m},\end{align}  where we recall that $ \mathbb{P}_ {\AAA^\intercal}={V}_\AAA{V}_\AAA^\intercal$ and consider  the projector 
\begin{align}\label{eq:hatP}
\mathbb{P}_{\ZZZ,k}={U}_{\ZZZ,k} {{U}_{\ZZZ,k}}^\intercal.
\end{align}
Matrix \eqref{eq:hatP} appears in the closed-form solution of~\eqref{eq:prob1}, as shown in  the following  theorem. The proof  is given in \cite{HeasHerzet2021}.

\noindent
\begin{theorem}\label{prop22}
Problem~\eqref{eq:prob1}  admits the following solution 
\begin{align}\label{eq:Sol}
A_k^\star= \mathbb{P}_{\ZZZ,k}   \BBB \AAA^{\dagger}.
\end{align} 
Moreover, the optimal approximation error can be expressed as
\begin{align}\label{eq:errorEstim}
 \|\BBB -A_k^\star \AAA \|^2_F = \sum_{i=k+1}^m \sigma_{\ZZZ,i}^2 +\| \BBB (I_m-\mathbb{P}_{\AAA^{\intercal}})\|_F^2.
\end{align}
  \end{theorem}

 In words, Theorem~\ref{prop22} shows that problem~\eqref{eq:prob1} is simply solved by computing the orthogonal projection of  the  solution of the unconstrained  problem~\eqref{eq:prob1_unconst},  onto the subspace spanned by the first $k$    left singular vectors  of  $ \ZZZ.$    The $\ell_2$-norm of the error is simply expressed in terms of  the singular values of $\ZZZ$, and  the square norm of the projection of the rows of $\BBB$ onto the orthogonal of the image of $\AAA^\intercal$.
If $\AAA$ is full {row-}rank,  we then obtain  the simplifications $\mathbb{P}_{\AAA^\intercal}=I_m$ and $\ZZZ=\BBB$. In this case, the second term in the right-hand side of~\eqref{eq:errorEstim} vanishes and the  approximation error reduces to $ \|\BBB -A_k^\star \AAA \|^2_F = \sum_{i=k+1}^m \sigma_{\BBB,i}^2$. The latter error is independent of matrix $\AAA$ and is simply the sum of the square of the $m-k$ smallest singular values of $\BBB$. This error also corresponds to the optimal error for the approximation $\BBB$ by a matrix of rank at most $k$ in the Frobenius norm \cite{eckart1936approximation}.  

  Besides, note that $r=\textrm{rank}(A_k^\star)$  can be smaller than $k$. Indeed, by the Sylvester's theorem \cite{Horn12} we have  that
\begin{align*}
r&\le\min(\textrm{rank}(\mathbb{P}_{\ZZZ,k}  ), \textrm{rank}(\BBB\XXX^\dagger))  \le  \textrm{rank}(\BBB\XXX^\dagger)\\
&\le \min(\textrm{rank}(\BBB), \textrm{rank}(\AAA^\dagger))= \min(\textrm{rank}(\BBB), \textrm{rank}(\AAA)),
\end{align*} 
which shows that $r<k$ if $\textrm{rank}(\XXX)$ or $\textrm{rank}(\YYY)$ is smaller than $k$, but also if $\textrm{rank}(\BBB\XXX^\dagger) < k$. 


It is worth mentioning that  a generalization of Theorem~\ref{prop22} to separable infinite-dimensional Hilbert spaces is  proposed in~\cite{HeasHerzet18Maps}. This generalization  characterizes the solution of  low-rank approximations in reproducing kernel Hilbert spaces (where $n=\infty$) at the core of  kernel-based DMD \cite{HeasIcassp2020,williams2014kernel}, and characterizes the solution of the DMD counterpart (where $m=\infty$) to  the continuous POD problem presented in \cite[Theorem 6.2]{quarteroni2015reduced}.

\subsection{Algorithm Evaluating $A_k^\star$}

\begin{algorithm}[!h]
\begin{algorithmic}[0]
\State \textbf{inputs}: $(\XXX,\YYY).$
\State 1)  Compute the SVD of $\AAA= V_{\AAA} \Sigma^\dagger_{\AAA} U_{\AAA}^\intercal  $ 
\State 2)  Compute $\mathbf{Z}=\BBB V_{\AAA}\Sigma_{\AAA}\Sigma_{\AAA}^\dagger V_{\AAA}^\intercal$.
\State 3) Compute the SVD of $\mathbf{Z}$ to obtain the projector $\mathbb{P}_{\ZZZ,k}$.
\State 4) Compute $A_k^\star= \mathbb{P}_{\ZZZ,k}   \BBB  V_{\AAA} \Sigma^\dagger_{\AAA} U_{\AAA}^\intercal  $.
\State \textbf{output}: $A_k^\star$. 
\end{algorithmic}
\caption{Computation of $A_k^\star$, a solution of~\eqref{eq:prob1} \label{algo:1}}
\end{algorithm}
The design of an algorithm computing   the solution~\eqref{eq:Sol} is straightforward: evaluating $A_k^\star$ consists in making a product of easily-computable matrices.   The proposed procedure is summarized in Algorithm~\ref{algo:1}.

Steps 1) to 3) of  Algorithm~\ref{algo:1} implies the computation of the SVD of  matrices $\AAA,\ZZZ \in \Rr^{n \times m }$, and matrix multiplications involving $m^2$ vector products in $\Rr^n$ or $\Rr^m$. 
The  complexity  of these  first three  steps is therefore  $\mathcal{O}(m^2(m+n))$. 
 Computing explicitly each entry of  $A_k^\star \in \Rr^{n\times n}$ in  step 4) of Algorithm~\ref{algo:1}  then requires a  complexity of   $\mathcal{O}(n^2k)$, which is prohibitive for large $n$. However, as detailed in the next section, this last step is not necessary to   factorize the optimal solution  $A^\star_k$ in the form of \eqref{eq:solFactor} or \eqref{eq:factorEVD}.  \vspace{-0.15cm}



\subsection{Algorithms Factorizing $A_k^\star$}
Given the closed-form solution~\eqref{eq:Sol}, we  present in what follows how to compute from $\AAA$ and $\BBB$ a factorization of the optimal solution  $A^\star_k$ in the form of \eqref{eq:solFactor} or \eqref{eq:factorEVD}. We will need 
  matrix
\begin{align}\label{eq:defhatQ}
 {W} =({{U}_{\ZZZ,k}}^\intercal  \BBB \AAA^\dagger)^\intercal \in \Rr^{n \times k}.
 \end{align} \vspace{-0.15cm}

\textbf{Factorization of the form of \eqref{eq:solFactor}.} 
By performing the first three steps  of Algorithm~\ref{algo:1} and then making the identifications  $\R={U}_{\ZZZ,k}$ and $\Q={W}$, we obtain a factorization of $A_k^\star$ of the form of \eqref{eq:solFactor}. As mentioned in the introduction, trajectories of \eqref{eq:model_koopman_approx} can then be computed with system~\eqref{eq:genericROM0}  setting $R={U}_{\ZZZ,k}$, $L=W$ and $S=  W^\intercal  {U}_{\ZZZ,k}$.
The method relies on the  first three steps  of Algorithm~\ref{algo:1} and on the computation of matrix ${W} $. The three steps in Algorithm~\ref{algo:1} imply a complexity of $\mathcal{O}(m^2(m+n))$ while the  computation of ${W} $ requires a complexity of $\mathcal{O}(n k^2)$. Since $k \le m$, the off-line complexity to build  the  factorization~\eqref{eq:solFactor}  from  $\AAA$ and $\BBB$ scales as $\mathcal{O}(m^2(m+n))$, which is the same order of complexity as the procedures described in Section~\ref{sec:stateArt}.  \\

\begin{algorithm}[t]
\begin{algorithmic}[0]
\State \textbf{inputs}:  $(\XXX,\YYY).$
\State 1) Compute step 1 to 3 of Algorithm~\ref{algo:1} and  use~\eqref{eq:defhatQ} to obtain ${W}$.
\State 2) Let $r=\textrm{rank}(A^\star_k)$ and solve for $ i=1, \ldots,r$ the eigen-equations $$ ({W}^\intercal {U}_{\ZZZ,k}) w^r_i = \lambda_i w^r_i\quad \textrm{and}\quad ({{U}_{\ZZZ,k}}^\intercal {W}) w^\ell_i =  \lambda_i w^\ell_i,$$ where  $w^r_i, w^\ell_i \in \Cr^k$  
and $\lambda_i\in \Cr$ such that $|\lambda_{i+1}| \ge |\lambda_i |$. 

\State 3) Compute for $i=1,\ldots,r$ the right  and  left eigenvectors  
\begin{align}\label{eq:defEigenvectors}
  \zeta_i=  {U}_{\ZZZ,k}   w^r_i\quad \textrm{and}\quad   {\xi_i}=  {W} w^\ell_i.
  \end{align} 
  \State 4) Rescale the ${\xi_i}$'s so that $  {\xi_i}^T \zeta_i =1$. 
\State \textbf{outputs}:   $L=(\xi_1\cdots \xi_r)$,  $R=(\zeta_1\cdots \zeta_r)$,    $S=\textrm{diag}(\lambda_1,\cdots, \lambda_r).$ 
\end{algorithmic}
\caption{EVD of $A^\star_k$ or low-rank DMD  \label{algo:2}}
\end{algorithm}

\textbf{Factorization of the form of  \eqref{eq:factorEVD}.}\label{sec:EVDAstar} 
According to the previous factorization of the form of \eqref{eq:solFactor}, $A^\star_k$ is  the product of matrix ${U}_{\ZZZ,k}$ in $\Rr^{n \times k}$ with matrix ${W}^\intercal$ in $\Rr^{k \times n}$. Therefore, using  standard matrix analysis, we expect the eigenvectors of $A^\star_k$ to belong to a $k$-dimensional  subspace  \cite{golub2013matrix}. As shown in  the next proposition,  the non-zero eigenvalues of $A^\star_k$  are obtained by EVD of certain matrices in $\Rr^{k \times k}$. The proof of this proposition  is given in \cite{HeasHerzet2021}.

 \begin{proposition}\label{rem:2}
 Assume $A^\star_k$ is diagonalizable.   The elements of  $\{\zeta_i, \xi_i, \lambda_i\}_{i=1}^{\mathrm{rank}(A^\star_k)}$ generated  by Algorithm~\ref{algo:2}  are  the  right eigenvectors, the left eigenvectors and the eigenvalues  of the economy size EVD of  $A^\star_k$. 
 \end{proposition}

In words, Proposition~\ref{rem:2} shows that Algorithm~\ref{algo:2} computes  the  EVD   of  $A^\star_k$ by  diagonalizing  two  matrices in  $\Rr^{k\times k}$. 
 The complexity  to build  this  EVD  from snapshots $\XXX$ and $\YYY$ is  $\mathcal{O}(m^2(m+n))$. More precisely, as mentioned previously, performing the first three  steps of Algorithm~\ref{algo:1} (\ie step 1) of Algorithm~\ref{algo:2}) requires a number of operations scaling as $\mathcal{O}(m^2(m+n))$;
  the complexity of step 2) is $\mathcal{O}(k^3)$ since it  performs the EVDs of  $k \times k$ matrices; 
  step 3) involves $r\times n$ vector products in $\Rr^m$ while step 4) involves $r$ vector products in $\Rr^n$, with $r \le k \le m$. 
  {Overall, the complexity of Algorithm~\ref{algo:2} is dominated by step 1) and the EVD of $A^\star_k$ can be evaluated with a computational cost of the order of $\mathcal{O}(m^2(m+n))$.}\vspace{-0.25cm}

  \section{Conclusion}

This work reviews the state-of-the-art algorithms proposed to compute low-rank DMD. In particular, it details an exact solution to this problem  computable  with a complexity of  the same order as  state-of-the-art sub-optimal methods.

\appendix

\section{Analytical Comparison between Low-Rank DMD and TLS DMD }\label{app:comp}

 This simple example demonstrates that the solution given by TLS-DMD can be biased on the contrary to the one given by the optimal algorithm  for low-rank DMD. It also shows that in the case (favorable to TLS-DMD) where the solution given by TLS-DMD is unbiased, the noise robustness of the two approaches are comparable. 
	
	\begin{example}

Consider the case where $k=1$, the data $\XXX$ is set either to $$\AAA_1=\begin{pmatrix}1 &0\\0 &10\\1 &10\end{pmatrix}\quad \textrm{or}\quad \AAA_2=\begin{pmatrix}1 &0\\0 &1\\1 &1\end{pmatrix},$$ and
   $$\YYY=\begin{pmatrix}5 &0\\\epsilon &2\\10 &0\end{pmatrix},$$ where $\epsilon$ is a scalar representing a small perturbation of the matrix entry.   
   
   \begin{itemize}
	\item {\bf Low-rank DMD solution.}
Since in both cases $\XXX$  is  full rank, we have $\ZZZ=\YYY$. Straightforward analytical calculations yield the vector ${U}_{\ZZZ,1}=\frac{(\sigma_2-4)}{(\sigma_2((\sigma_2-4)^2+4\epsilon^2))^{1/2}}\begin{pmatrix}5\\\epsilon \\10 \end{pmatrix},$ with the singular value $\sigma_2=\frac{129+\epsilon+\sqrt{14641+240\epsilon +17 \epsilon^2}}{2}$.    This leads to an optimal rank-$1$ solution (given by our algorithm)
\begin{align*}
A_1^\star=\mathbb{P}_{\ZZZ,1}   \BBB \AAA^{\dagger}&=\frac{125(\sigma_2-4)^2}{\sigma_2((\sigma_2-4)^2+4\epsilon^2)}\begin{pmatrix}5+0.04\epsilon^2 &0.08\epsilon\\\epsilon+0.008\epsilon^3 &0.016\epsilon^2\\10 +0.08\epsilon^2&0.16\epsilon\end{pmatrix}\AAA^{\dagger}.
\end{align*}
   The non-zero eigen value $\lambda(A_1^\star)$ of matrix $A_1^\star$ corresponds to the solution of a linear equation. It takes the form of
     \begin{align*}
\lambda(A_1^\star)&=\frac{125(\sigma_2-4)^2}{\sigma_2((\sigma_2-4)^2+4\epsilon^2)}\left(\frac{20}{3}-a \epsilon+ b\epsilon^2-0.008\epsilon^3\right),
\end{align*}
   with $a=0.992$ (resp. $a=0.920$), $b=0.0544$ (resp. $b=0.0540$) for $\AAA=\AAA_1$ (resp. $\AAA=\AAA_2$). For a small perturbation $\epsilon$, we obtain for $\AAA=\AAA_1$ the approximation $$\lambda(A_1^\star)
\approx \frac{20}{3}-0.3416\epsilon,$$ 
(resp. 
  $\lambda(A_1^\star)
\approx \frac{20}{3}-0.3168\epsilon$ for $\AAA=\AAA_2$).
For $\epsilon=0$, the optimal error norm is
  $$\| \BBB-A_1^\star \AAA \|_F=2.00,$$  equally for $\XXX=\XXX_1$ and $\XXX=\XXX_2$.\\

	\item {\bf TLS-DMD  solution.}
  We remark that  $\mathbf{K}^\intercal \mathbf{K}=\begin{pmatrix}127+ \epsilon^2 &10+2\epsilon\\10+2\epsilon &204\end{pmatrix}$ for $\AAA=\AAA_1$ (resp.   $\mathbf{K}^\intercal \mathbf{K}=\begin{pmatrix}127+ \epsilon^2 &1+2\epsilon\\1+2\epsilon &10\end{pmatrix}$ for $\AAA=\AAA_2$) and simple algebra yields the singular vector
   $${V}_{\mathbf{K},1}=\frac{1}{\|\begin{pmatrix}10 +2\epsilon\\ \sigma_1 -127\end{pmatrix} \|_2}\begin{pmatrix}10 +2\epsilon\\ \sigma_1 -127\end{pmatrix}, \quad\textrm{with}\quad \sigma_1=\frac{331+\sqrt{ 6329 +160 \epsilon-138 \epsilon^2}}{2}+\mathcal{O}(\epsilon^2),$$
   $$\left(\textrm{resp. }{V}_{\mathbf{K},1}=\frac{1}{\|\begin{pmatrix}1 +2\epsilon\\ \sigma_1 -127\end{pmatrix} \|_2}\begin{pmatrix}1 +2\epsilon\\ \sigma_1 -127\end{pmatrix}, \quad\textrm{with}\quad \sigma_1=\frac{137+\sqrt{ 13693 +16 \epsilon-250 \epsilon^2}}{2}+\mathcal{O}(\epsilon^2).\right)$$
The TLS-DMD  solution  provided by {\it Hemati et al.} is
\begin{align*}
\hat A_1&=
\BBB
{V}_{\mathbf{K},1}{{V}_{\mathbf{K},1}}^\intercal
\AAA^{\dagger},
  \end{align*}
  and the non-zero eigen value $\lambda(\hat A_1)$ of matrix $\hat A_1$ is the solution of a linear equation, more explicitly   for $\AAA=\AAA_1$ (resp. $\AAA=\AAA_2$) 
      \begin{align*}
\lambda(\hat A_1)&=\frac{20+(\sigma_1-127)(a(\sigma_1-127)-b)}{3+0.03(\sigma_1-127)^2}+ \mathcal{O}(\epsilon),
\end{align*}
with the constants $a=0.004$ (resp. $a=0.04$) and $b=0.15$ (resp. $b=0.30$). We obtain    for $\AAA=\AAA_1$  the approximation
$$\lambda(\hat A_1) \approx 0.1754+   0.0312\epsilon,$$
 (resp. $\lambda(\hat A_1) \approx 6.6746 -   0.3127\epsilon$    for $\AAA=\AAA_2$). For $\epsilon=0$, the Frobenius error norm related to {\it Hemati et al.}'s solution is  $$\| \BBB-\hat A_1 \AAA \|_F\approx 11.09$$ for $\XXX=\XXX_1$ (resp.   $\| \BBB-\hat A_1 \AAA \|_F\approx 2.0021$ for $\XXX=\XXX_2$).\\

We note that in the case where $\XXX=\XXX_2$, the eigenvalue estimated with the approach of {\it Hemati et al.} is almost equal to the proposed eigenvalue estimate, whereas it is strongly biased for $\XXX=\XXX_1$. Moreover, in the favourable case where $\XXX=\XXX_2$, a small perturbation on the matrix input induces similar biases for the TLS-DMD and the low-rank DMD. 
 \end{itemize}

	\end{example}

\bibliographystyle{spmpsci}      
\bibliography{./bibtex}
\end{document}